\documentclass[letterpaper, 10 pt, journal, twoside]{IEEEtran}
\usepackage{cite}
\usepackage{amsmath}
\usepackage{graphicx}
\usepackage{float} 
\usepackage{subfigure}
\usepackage{overpic}
\usepackage{wrapfig}
\usepackage{balance} 
\usepackage{ulem}
\usepackage{color}
\usepackage[colorlinks,linkcolor=red]{hyperref} 
\usepackage{multirow}
\usepackage{mathtools}
\usepackage{booktabs} 

\IEEEoverridecommandlockouts                              

\begin{document}

\title{Active Implicit Object Reconstruction using \\
Uncertainty-guided Next-Best-View Optimization
}

\author{Dongyu Yan*$^{1}$, Jianheng Liu*$^{1}$, Fengyu Quan$^{1}$, Haoyao Chen$^{1}$  and Mengmeng Fu$^{2}$
\thanks{*D.Y. Yan and J.H. Liu contributed equally.}

\thanks{Manuscript received: March, 29, 2023; Revised: July, 2, 2023; Accepted: July, 29, 2023.
This paper was recommended for publication by Editor Sven Behnke upon evaluation of the Associate Editor and Reviewers' comments.
This work was supported in part by the National Natural Science Foundation of China under Grants U21A20119 and U1713206 and in part by the Shenzhen Science and Innovation Committee under Grants JCYJ20200109113412326 and JCYJ20210324120400003.
(Corresponding author: Haoyao Chen and Mengmeng Fu.)}%

\thanks{D.Y. Yan, J.H. Liu, F.Y. Quan and H.Y. Chen are with the School of Mechanical Engineering and Automation, Harbin Institute of Technology Shenzhen, P.R. China.
  {\tt\footnotesize \{21s053072, liujianheng, 18b353011\}@stu.hit.edu.cn, hychen5@hit.edu.cn}.}
\thanks{M.M. Fu is with Department of Neurosurgery, Shenzhen University General Hospital, Shenzhen, P.R. China.
  {\tt\footnotesize fumengmeng2019@163.com}.}%
\thanks{Digital Object Identifier (DOI): see top of this page.}
}%

\markboth{IEEE Robotics and Automation Letters. Preprint Version. Accepted JULY, 2023}
{Yan \MakeLowercase{\textit{et al.}}: Active Implicit Object Reconstruction using Uncertainty-guided Next-Best-View Optimization} 

\maketitle

\begin{abstract}

Actively planning sensor views during object reconstruction is crucial for autonomous mobile robots.
An effective method should be able to strike a balance between accuracy and efficiency.
In this paper, we propose a seamless integration of the emerging implicit representation with the active reconstruction task.
We build an implicit occupancy field as our geometry proxy.
While training, the prior object bounding box is utilized as auxiliary information to generate clean and detailed reconstructions.
To evaluate view uncertainty, we employ a sampling-based approach that directly extracts entropy from the reconstructed occupancy probability field as our measure of view information gain.
This eliminates the need for additional uncertainty maps or learning.
Unlike previous methods that compare view uncertainty within a finite set of candidates, we aim to find the next-best-view (NBV) on a continuous manifold.
Leveraging the differentiability of the implicit representation, the NBV can be optimized directly by maximizing the view uncertainty using gradient descent.
It significantly enhances the method's adaptability to different scenarios.
Simulation and real-world experiments demonstrate that our approach effectively improves reconstruction accuracy and efficiency of view planning in active reconstruction tasks.


\end{abstract}

\section{Introduction}

\IEEEPARstart{A}{ctive} reconstruction is important in various autonomous robotics tasks like perception, manipulation, and scene understanding.
The primary objective is to enable the robot to reconstruct the finest model using the fewest views.
This requires the autonomous system to evaluate current scene information, plan for future movements, and integrate the newly acquired information into the map.
The most widely adopted approach is NBV planning, where at each step, the system selects the most uncertain view that brings the greatest information gain as the next target location.

\begin{figure}[t]
    \centering
    \includegraphics[width = 0.48\textwidth]{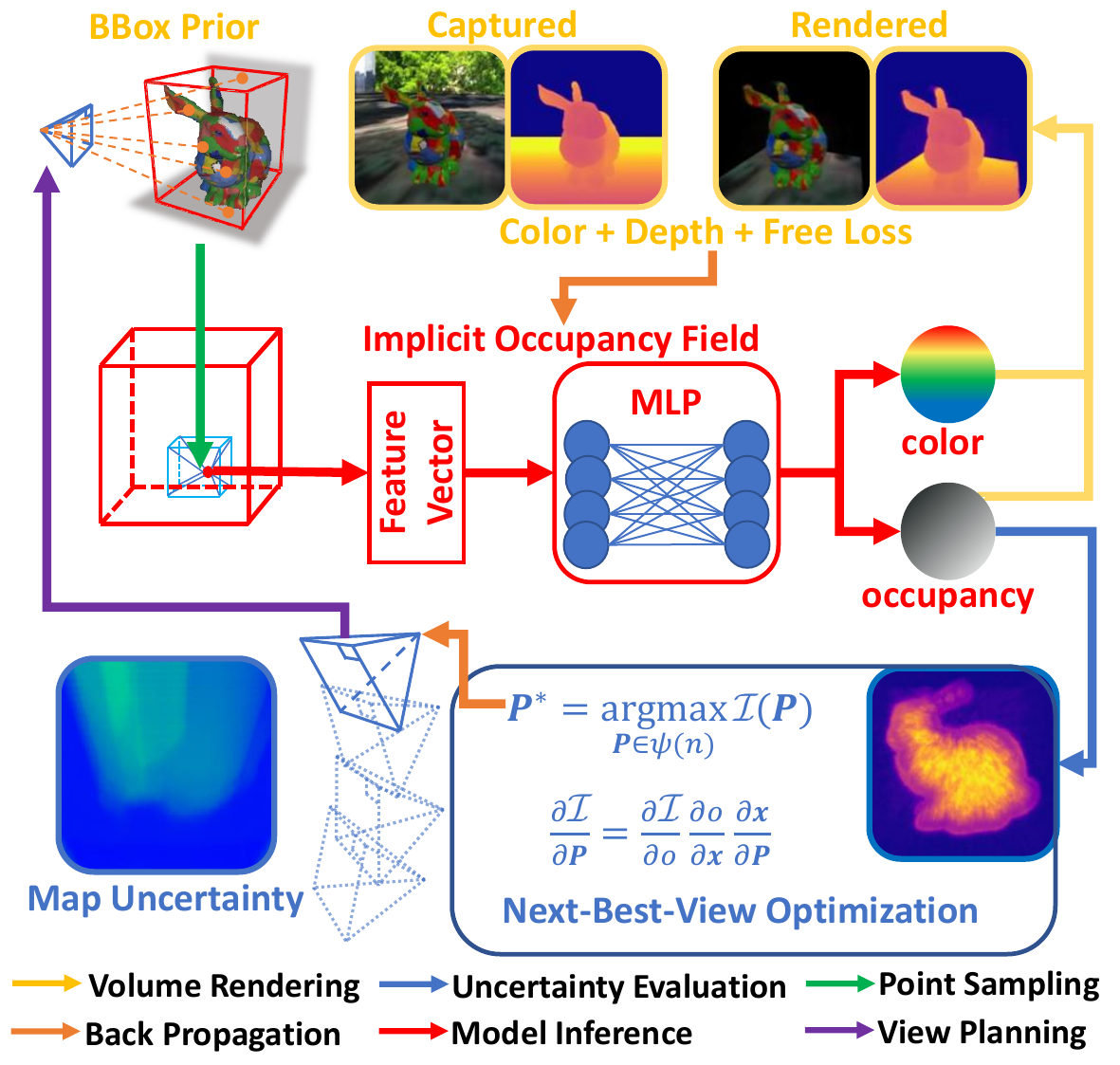}
    \caption{
        The architecture of our method.
        We construct our implicit occupancy field from volume-rendered color and depth supervision and additional free-ray supervision.
        Object bounding box prior is utilized to assist sampling and model range.
        We evaluate view uncertainty directly from occupancy probability using a sampling-based approach.
        The NBV pose is iteratively optimized by maximizing the information gain through back propagation and is used to guide the robot movement for next cycle reconstruction.
        }
    \label{fig:flow}
    \vspace{-4mm}
\end{figure}

In traditional NBV methods, uncertainty information is typically stored in the form of voxel representation \cite{isler2016information}.
Evaluating information gain for each view entails traversing all voxels in the map to determine their visibility and occlusion.
This traversal-based method is inefficient and results in slow evaluation speed.
Additionally, the discrete nature of uncertainty maps can lead to non-optimal view selection, reducing the overall performance of view planning.

With the recent advancements in NeRFs \cite{mildenhall2020nerf, sitzmann2020implicit}, 3D object reconstruction using implicit representation has demonstrated significant advantages in differentiability, high-precision reconstruction, and low memory footprint.
To achieve autonomous reconstruction using implicit representation, methods have explored to perform active view planning using implicit representation \cite{lee2022uncertainty, ran2023neurar, jin2023neu, pan2022activenerf}.
They have showcased high accuracy in reconstruction and novel view synthesis but still encounter challenges.
Methods with color-only inputs \cite{lee2022uncertainty, jin2023neu, pan2022activenerf} require model initialization with predefined reference views before the autonomous process starts.
The reference views are also needed by methods using learning-based uncertainty \cite{ran2023neurar, jin2023neu, pan2022activenerf}, otherwise, the uncertainty prediction of the unknown part is not trained and may generate unwanted results.
Moreover, adhering to traditional candidate view settings limits their potential for reaching the optimal solution.
Additionally, the slow training speed of networks poses practical challenges when applying these methods to robotic systems.

In this work, we address the above problems by introducing our active implicit object reconstruction using uncertainty-guided NBV optimization.
Fig.~\ref{fig:flow} shows the architecture and pipeline of our method.
We focus on object-level reconstruction and use the implicit occupancy field as the scene representation. It can guarantee precise geometry and provide evaluation metrics for the NBV selection.

To enable our method to be used in real-time robotics applications, we follow the network architecture used in Instant-NGP \cite{muller2022instant}.
Depth information is also incorporated for faster convergence and fine-grained geometry.
By utilizing the bounded nature of objects, we limit our implicit function to a bounding box for full usage of the model capacity.
The bounding box also guides our sampling strategy during supervision, leading to more refined and cleaner object models.
Both free and valid rays provide sufficient information for uncertainty evaluation and high-precision reconstruction.

As for view uncertainty evaluation, we follow the occlusion-aware entropy metrics proposed in \cite{isler2016information}.
Since our method leverages the reconstructed occupancy probability field as the object geometry proxy, it can also be used for uncertainty evaluation.
We use a sampling-based method to evaluate per-view information gain.
Fast and effective evaluation can be achieved by directly sampling occupancy probability in the field of view and calculating the occlusion-aware entropy.
Additionally, we propose a top-N strategy to adjust local and global attention for detail reconstruction.

After establishing the evaluation strategy, the next step is to calculate the pose of the NBV to plan the robot's movement.
Based on the differentiability of the implicit representation, our proposed sampling-based information gain can be treated as a differentiable function with respect to the view pose.
In this way, we can directly optimize the NBV and do away with the traditionally used candidate view selection manner.
By guiding selection with optimization, the NBV can be estimated from a continuous manifold, which has higher adaptability to various scenarios without pre-defining a finite set of candidates.

In summary, our main contributions are the following:
\begin{itemize}
    \item We integrate active view planning with implicit reconstruction by constructing a unified occupancy probability map and using a sampling-based method for uncertainty evaluation.
    Our method significantly improves reconstruction completeness from baseline methods by finding more informative views with limited measurements.
    \item We propose a novel optimization-based NBV planning on a continuous manifold, demonstrating superior performance compared to conventional finite candidate view sets in terms of adaptability.
    \item We eliminate the additional pre- and post-processing steps associated with other implicit methods, thereby enabling a fully autonomous reconstruction process.
    Our method is real-time capable and can be applied to real-world robotic platforms.
\end{itemize}

\section{Related Work}

\subsection{Implicit Neural Reconstruction}

Representing geometry, appearance, and other scene attributes (e.g., surface normal, semantic label, etc.) with an implicit neural network has received much attention\cite{sitzmann2020implicit, mildenhall2020nerf}.
In contrast to traditional 3D representations, the implicit model represents scenes with a continuous function.
By giving a coordinate, the forward pass of the function returns scene information at the position.
The nature of implicit representation results in higher precision surface reconstruction with less memory footprint than traditional discrete representations.

To supervise the implicit function, neural radiance field (NeRF) \cite{mildenhall2020nerf} proposes using volumetric rendering to generate comparisons with the captured images.
Although good-looking synthesized views can be generated, the surface geometry lacks accuracy, mainly due to the use of volume density as its geometry attribute.
To this end, SDF\cite{wang2021neus, yariv2021volume} or occupancy\cite{niemeyer2020differentiable, oechsle2021unisurf} is introduced into volume rendering to attain better-generated geometry.
However, applying this to robotics applications is still hard due to its low training speed.
To resolve this drawback, depth information is fused\cite{azinovic2021neural, deng2021depth} to help the model converge faster and solve the ambiguity of geometry and appearance.
Some methods\cite{fridovich2022plenoxels, sun2022direct, muller2022instant} utilize voxel-grids to store features discretely and use interpolation and a shallow decoder to infer the final result.
This can also boost training speed, due to the simple computation and centralized learnable features.

Our reconstruction method follows the aforementioned depth-aided supervision and feature grid structure to achieve real-time ability in our robotics tasks.
Specifically, we utilize the prior knowledge of the object bounding box along with the free space supervision and propose a novel reconstruction pipeline for object-level reconstruction.

\subsection{NBV Planning}

NBV planning aims to obtain the next sensor view that can best reduce the reconstructed object's uncertainty.
According to different uncertainty representations, traditional methods can be divided into frontier-based\cite{yamauchi1997frontier, vasquez2014volumetric}, surface-based\cite{pito1999solution, chen2005vision, wu2014quality} and volumetric-based\cite{isler2016information, daudelin2017adaptable} methods.
Frontier-based methods take voxel-grids as the map representation and define the border of free and unknown voxels as frontiers\cite{yamauchi1997frontier}.
Then, the NBV is further determined using the frontier information.
Surface-based methods represent the global map as a 3D surface model, such as mesh, surface points, and signed distance function.
Once the surface model is generated, the algorithm can follow its contour or curvature tendency to guide NBV selection\cite{wu2014quality}.
Unlike these approaches, volumetric-based methods\cite{isler2016information, daudelin2017adaptable} analyze the full spatial information of the view.
They take the traversed voxels by ray marching to calculate NBV metrics, and the view with the highest information gain is selected from among all candidate views as the next target position of the robot.

With recent developments of the neural radiance field, some methods incorporate the benefits of neural rendering and implicit representations into the active reconstruction process \cite{lee2022uncertainty, pan2022activenerf, jin2023neu, ran2023neurar}.
Reference \cite{lee2022uncertainty} represents object implicitly by color and density.
It performs volume rendering of a novel view and uses the weight function along rays to measure the uncertainty.
ActiveNeRF \cite{pan2022activenerf} models the scene's color as Gaussian distribution and leverages the network to learn it.
The uncertainty is then evaluated by aggregating the variance within the field of view.
NeU-NBV \cite{jin2023neu} learns a mapless, image-based neural rendering framework and predicts pixel uncertainty from nearby reference views through the pre-trained network.
These methods obtain good reconstruction quality and uncertainty measurement.
However, using color-only input requires additional views for initialization before active reconstruction, and the reliance on predefined candidate views reduces adaptability to different scenarios.

Our approach uses a sampling-based method and leverages the reconstructed implicit occupancy probability for uncertainty evaluation.
Using the RGB-D image as input, our method can perform active reconstruction from start to finish.
To do away with the traditional candidate view selection manner, we optimize the NBV utilizing the differentiability of the implicit representation.

\section{Implicit Reconstruction}
\label{sec:recon}

\subsection{Background}\label{sec:background}

Our implicit model can be represented as:
\begin{equation}
o(\mathbf{x}),\mathbf{c}(\mathbf{x})=f(\mathbf{x},\Theta),
\end{equation}
where $\mathbf{x}$ represents the input coordinates, and $\Theta$ is the network parameter that implicitly saves the scene information.
The output $o(\mathbf{x}) \in [0, 1]$ represents the occupancy probability and $\mathbf{c}$ is the color.
We follow the reconstruction pipeline of UNISURF \cite{oechsle2021unisurf} to build our implicit occupancy field and additionally add depth into supervision.
Given the camera intrinsic $\mathbf{K}$ and extrinsic $\left\{\mathbf{P}_i\right\}_{i=1}^{N_{v}}$  of $N_v$ views, we sample $N_s$ points on the rays cast by each pixels.
Then, we calculate the color and depth of these pixels by volume rendering:
\begin{equation}
    \hat{\mathbf{C}}(\mathbf{r})=\sum_{i=1}^{N_s} w\left(\mathbf{x}_i\right)\mathbf{c}\left(\mathbf{x}_i\right),
    \hat{D}(\mathbf{r})=\sum_{i=1}^{N_s}
    w\left(\mathbf{x}_i\right)d\left(\mathbf{x}_i\right),
    \label{eq:render}
\end{equation}
where, $\mathbf{x}_i$ denotes the $i^{th}$ sampled points on ray $\mathbf{r}$ and $d(\mathbf{x}_i)$ denotes the depth value at position $\mathbf{x}_i$.
The weight $w\left(\mathbf{x}_i\right)$ is the product of the occupancy value $o(\mathbf{x}_i)$ and the occlusion term $T\left(\mathbf{x}_i\right)$:
\begin{equation}
    w\left(\mathbf{x}_i\right)=o\left(\mathbf{x}_i\right)T\left(\mathbf{x}_i\right),
    T\left(\mathbf{x}_i\right)=\prod_{j=1}^{i-1}\left(1-o\left(\mathbf{x}_j\right)\right).
    \label{eq:weight}
\end{equation}
The supervision is added by comparing the ground truth color $\mathbf{C}(\mathbf{r})$ and depth $D(\mathbf{r})$ with the rendered ones:
\begin{equation}
\begin{aligned}
    \mathcal{L}_{color}&=\frac{1}{\left\|\mathcal{R}\right\|}\sum_{\mathbf{r} \in \mathcal{R}}\left\|\hat{\mathbf{C}}(\mathbf{r})-\mathbf{C}(\mathbf{r})\right\|_2^2, \\
    \mathcal{L}_{depth}&=\frac{1}{\left\|\mathcal{R}\right\|}\sum_{\mathbf{r} \in \mathcal{R}}\left\|\hat{D}(\mathbf{r})-D(\mathbf{r})\right\|_1,
    \label{eq:val}
\end{aligned}
\end{equation}
where, $\mathcal{R}$ denotes a batch of rays sampled on the given set of camera views.

Although this supervision method of implicit representation performs well in an ideal sensor setup and with dense input, it loses accuracy and generates floaters when these conditions are unmet.
Furthermore, without a region of interest to constrain it, the method may consider background information, which may affect NBV planning.
We further modify the implicit reconstruction pipeline specifically for object-level reconstruction.

\subsection{Implicit Supervision for Object}

Since we specify our reconstruction targets as objects, we can leverage the prior information of the axis-aligned bounding box (AABB) to further optimize the reconstruction process.

\subsubsection{Model Structure}

Inspired by Instant-NGP\cite{muller2022instant}, we utilize the hybrid representation of multi-resolution hash table and shallow multi-layer perceptron (MLP) as our model structure.
With the multi-resolution concatenation and the trilinear interpolation, high-frequency signals can be encoded efficiently.
By initializing all learnable parameters with mean 0 distribution, the occupancy domain can be initialized anywhere close to 0.5, which means that everywhere is unexplored with maximum uncertainty.
This initialization is essential to the later process of NBV planning.
This model structure significantly speeds up training, making our method real-time capable.

\subsubsection{Rays and Points Sampling}

\begin{figure}[htpb]
    \vspace{-4mm}
    \centering
    \includegraphics[width = 0.40\textwidth]{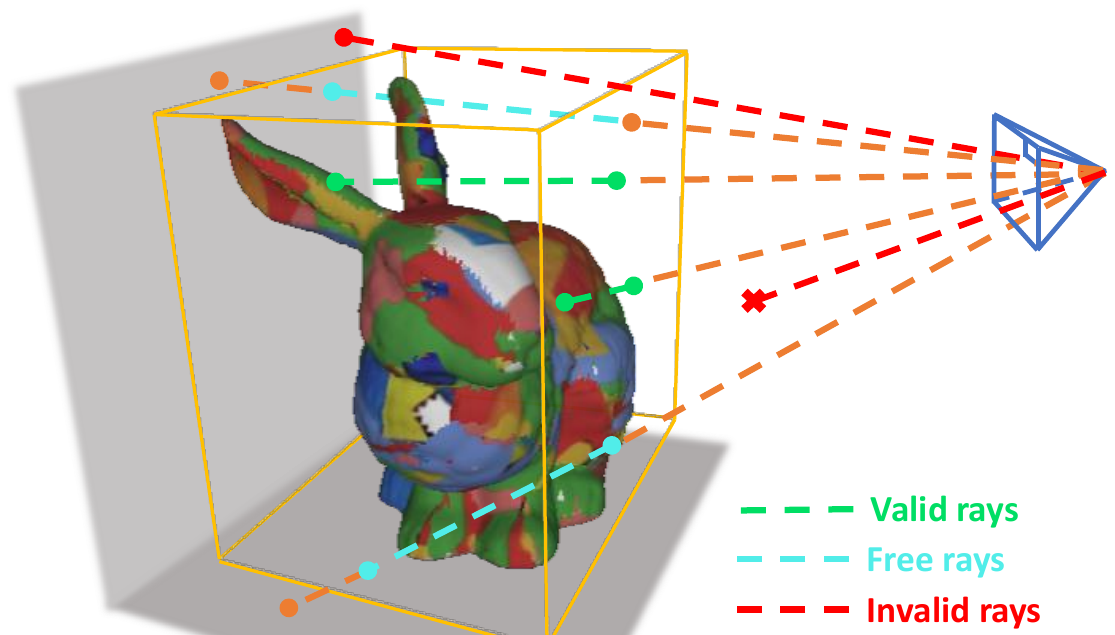}
    \caption{
      Illustration of ray types defined in our reconstruction method.
      The sampled rays are classified into invalid rays, valid rays and free rays.
    }
    \label{fig:rays}
\end{figure}

With the object bounding box, We can further guide the sampling and regularize the supervision of our implicit model.
Given an AABB $\{\mathbf{p}_{min},\mathbf{p}_{max}\}$ and $N_r$ rays uniformly sampled from the images, we first calculate their intersection by solving a system of linear equations.
Based on the intersection depth $\{d_{near}, d_{far}\}$ and the measured depth $d(\mathbf{r})$, we can classify the rays into 4 types, as shown in Fig. \ref{fig:rays}:
\begin{enumerate}
    \item{\makebox[5.6cm]{Rays with no intersection:\hfill} $d_{near}>d_{far}$.}
    \item{\makebox[5.6cm]{Rays with no depth measurement:\hfill} $d(\mathbf{r})=None$.}
    \item{\makebox[5.6cm]{Depth out of bounding box:\hfill} $d(\mathbf{r})>d_{far}$.}
    \item{\makebox[5.6cm]{Rays with valid depth measurement:\hfill} $Others$.}
\end{enumerate}

We treat 1) and 2) as invalid rays, as they do not contribute to the reconstruction.
We treat 3) as free rays, providing free space information in the bounding box.
We treat 4) as valid rays with good depth measurement, providing object occupancy information.

We constrain sample points using the AABB only to count useful information.
For free rays, we sample $N_f$ points in a stratified way.
For valid rays, we sample $N_s$ points near the estimated surface following a normal distribution $\mathcal{N}\left(d(\mathbf{r}),{\sigma_s}^2\right)$ which can focus more on surface regions and create a fine-grained model.

\subsubsection{Optimization}

In addition to the above-mentioned color and depth supervision (Eqn. \ref{eq:val}), we directly supervise points on free rays as non-occupied, using a binary-cross-entropy (BCE) loss:
\begin{equation}
    \mathcal{L}_{free}=\frac{1}{N_{f}}\sum_{k=0}^{N_{f}}-\log\left(1-o(\mathbf{x}_k)\right),
\end{equation}
where $N_{f}$ denotes the number of points sampled on the free ray.
The final loss function can be written as a linear combination of the above losses:
\begin{equation}
    \mathcal{L}=\mathcal{L}_{color}+\lambda_1 \mathcal{L}_{depth}+\lambda_2 \mathcal{L}_{free}.
    \label{eq:overall_loss}
\end{equation}

In practice, we find that the inaccuracy of robot pose estimation can cause misalignment of different views.
Therefore, we additionally add $\left\{\mathbf{P}_i\right\}_{i=1}^{N_{v}}$ into the optimization.
A more consistent result can be obtained through joint optimization of both the implicit object model and the sensor poses.

In general, our reconstruction approach considers both rays cast on the object and free space, creating full supervision of the bounded region.
This way, the view information can be fully merged into the global map, providing a good model understanding for NBV planning.

\section{NBV Planning}
\label{sec:NBV}

\subsection{View Uncertainty Evaluation}
\label{sec:view_eval}

The view uncertainty evaluation module aims to assign quantitative metrics to arbitrary views.
Since we use occupancy probability as our geometric proxy, we can extract uncertainty information directly from the reconstructed model without maintaining another uncertainty representation.
We follow the occlusion-aware volumetric information metrics introduced by \cite{isler2016information} but replace its voxel traversal by directly sampling occupancy probability in the field of view and calculating the occlusion-aware entropy.

We sample $N_p$ points on $N_e$ random rays cast by the given view, and the uncertainty of the sample point $\mathbf{x}$ can be represented by its entropy:
\begin{equation}
    \mathcal{I}_o(\mathbf{x})=-o(\mathbf{x}) \ln o(\mathbf{x})-\bar{o}(\mathbf{x}) \ln \bar{o}(\mathbf{x}).
\end{equation}
where $\bar{o}(\mathbf{x})$ is the complementary occupancy probability: $\bar{o}(\mathbf{x})=1-{o}(\mathbf{x})$.
The information considering occlusion takes the form of:
\begin{equation}
    \mathcal{I}_v(\mathbf{x})=T(\mathbf{x}) \mathcal{I}_o(\mathbf{x}),
\end{equation}
where $T(\mathbf{x})$ is the occlusion term mentioned in Eqn. \ref{eq:weight}.
We sum up the information of all sample points in the field of view to calculate the view uncertainty:
\begin{equation}
    \mathcal{I}=\frac{1}{N_e N_p}\sum_{i=1}^{N_e}\sum_{j=1}^{N_p} \mathcal{I}_v(\mathbf{x}_{ij}).
    \label{equ:sum}
\end{equation}

\begin{figure}[htpb]
    \vspace{-4mm}
    \centering
    \includegraphics[width=0.40\textwidth]{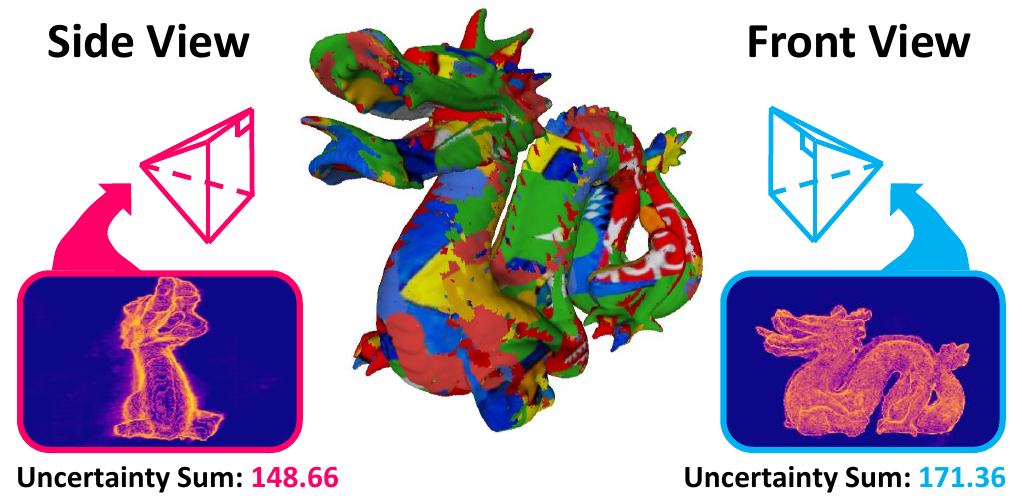}
    \caption{
        Illustration of the unfair evaluation in the dragon's scene.
        The bottom images show its uncertainty color map of different views after the $4^{th}$ reconstruction round.
        Although the front view (the right image) has already been well observed, the uncertainty sum is still significantly higher than that of the side view (the left image), which is not well reconstructed.
    }
    \vspace{-2mm}
    \label{fig:dragon}
\end{figure}

Although the above information gain can distinguish regions of uncertainty, there are still some problems preventing us from achieving fair and valuable evaluation:

1) The projected area of the target object differs in each view, resulting in an inherently unbalanced evaluation, as shown in Fig. \ref{fig:dragon}.

2) In the later period of the reconstruction process, the goal of active view planning should shift from building a general outline to focusing on details.

These problems are due to the use of summation to integrate the information of a view, which pays too much attention to the overall metrics and sacrifices the consideration of the details.
To this end, we propose a top-N evaluation criterion that can balance global and local information, formatted as:
\begin{equation}
    \label{eq:infomation}
    \mathcal{I}=\frac{1}{N_t N_p}\mathop{top}_{i=1}^{N_e}(\sum_{j=1}^{N_p} \mathcal{I}_v(\mathbf{x}_{ij}), N_t),
\end{equation}
where the function $\mathop{top}_{i=1}^{N_e}(v,N_t)$ entails choosing the top $N_t$ samples by their value $v$ from all $N_e$ candidates.
These criteria regularize the comparisons between different views.

Given our proposed modification of $\mathcal{I}$, we can obtain a trade-off for global or local attention by tuning the ratio of $N_t$ and $N_e$.
We set up a cosine scheduler to adjust the ratio according to the reconstruction process.
We dynamically decrease the selected ray number according to the total observed view number $N_v$ and gradually shift the attention from global information to local details, which is more adaptable to the reconstruction process.

\subsection{Optimization-based NBV Planning}

After the evaluation metrics of view uncertainty are defined, a well-designed strategy is needed to find the best NBV.
The most straightforward approach involves defining a set of candidate views in advance.
This way, we can determine the NBV by traversing the candidates and selecting the view with the largest uncertainty.
However, despite its advantages in simplicity, it may lead to sub-optimal view selection and affect reconstruction results.
The fixed distribution of candidate views may also cause low adaptability to different scenarios.

\begin{figure*}[t]
    \centering
    \begin{minipage}[b]{0.3\textwidth}
    \centering
        \includegraphics[width=\textwidth]{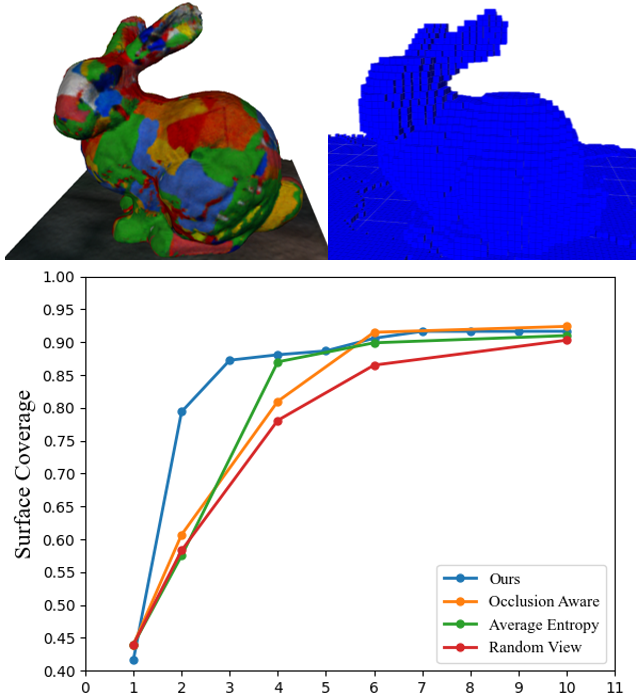}
        \footnotesize{(a) Stanford Bunny}
    \end{minipage}
    \begin{minipage}[b]{0.3\textwidth}
    \centering
        \includegraphics[width=\textwidth]{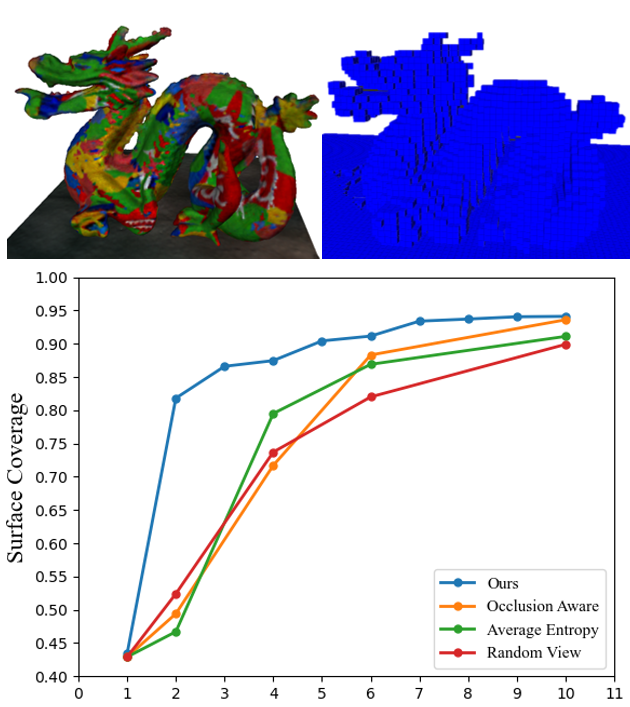}
        \footnotesize{(b) Dragon}
    \end{minipage}
    \begin{minipage}[b]{0.3\textwidth}
    \centering
        \includegraphics[width=\textwidth]{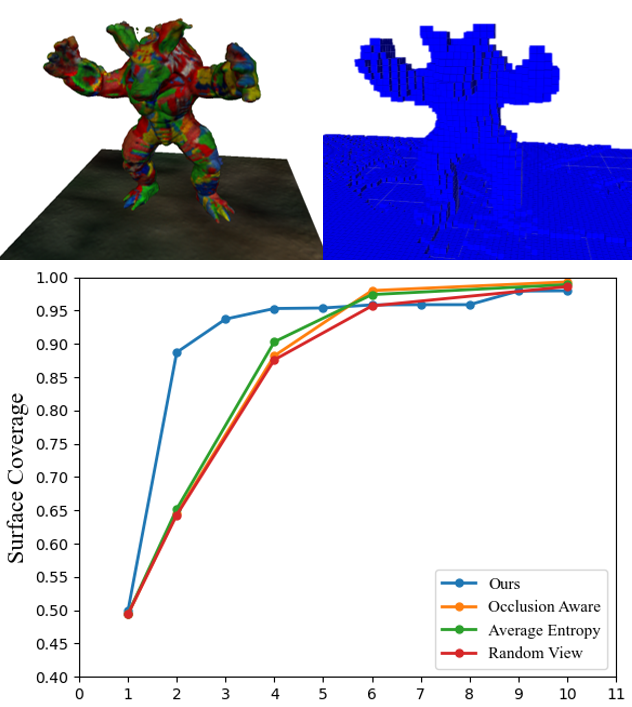}
        \footnotesize{(c) Armadillo}
    \end{minipage}
    \caption{
        Reconstructed model and surface coverage curve of Stanford Bunny, Dragon, and Armadillo.
        We compare our method against uncertainty policies of \textit{Occlusion-Aware}, \textit{Average-Energy} used in  \cite{isler2016information}, and \textit{Random View} on reconstruction quality and surface coverage.
        Our mesh extraction of the implicit surface has finer geometry and smoother surface than voxel representations.
        The surface coverage representing the reconstruction efficiency also converges faster.
    }
    \vspace{-2mm}
    \label{fig:compare_result}
\end{figure*}

In order to break through the limitations of selecting by candidate views and take more advantage of implicit representation, we manage to directly optimize the NBV $\mathbf{P}$ by maximizing the view uncertainty through gradient descent, leveraging the differentiability of the implicit object model.
By giving a novel view position, points are sampled under sensor coordinate. The transformation from sensor to world coordinate can be achieved by the sensor pose \textbf{P}. We then feed the points to the implicit function to obtain occupancy probability. Finally, the information gain can be acquired by our proposed policy. To this end, we have the information gain $\mathcal{I}$ as a differentiable function of the view pose $\textbf{P}$, written as $\mathcal{I}(\textbf{P}) \coloneqq \mathcal{I}$. The flow of gradient has the format of:
\begin{equation}
    \frac{\partial{\mathcal{I}}}{\partial{\mathbf{P}}}=
    \frac{\partial{\mathcal{I}}}{\partial{o}}
    \frac{\partial{o}}{\partial{\mathbf{x}}}
    \frac{\partial{\mathbf{x}}}{\partial{\mathbf{P}}}.
\end{equation}
By using the back-propagation method, we can solve the optimal view pose that maximizes the information through an optimization method:
\begin{equation}
    \mathbf{\tilde{P}}=\arg\max_{\mathbf{P}\in\mathit{\Psi}(n)}\mathcal{I}(\mathbf{P}),
\end{equation}
where $\mathit{\Psi}(n)$ ($n\leq6$) is a sub-manifold of $\mathit{SE}(3)$, which is used to limit the optimization space.
The optimization-based method does away with the non-optimal candidate views and achieves accurate selection without needing preset.

Nevertheless, in practice, the way to initialize the pose to be optimized after reconstruction still requires careful design.
The simple "Start from Last" strategy is prone to get stuck in a local maximum due to the great non-convexity of the objective function.

To better initialize the pose to be optimized, we developed an "Initialize by Sampling" method.
Before optimization, we first randomly sample $N_k$ views in the manifold $\mathit{\Psi}(n)$ and evaluate their uncertainty.
And then, we initialize the pose to the one with the largest uncertainty.
By using the "Initialize by Sampling" method, we can start optimizing from a point close to the global maximum.
This can avoid falling into the local maximum and simplify the optimization task.
The sampling density can be set sparsely since only a rough guide is needed.

Along with uncertainty metrics, task-oriented factors can be included in the objectives.
In order to avoid collision, we punish the occupancy value of the view position by adding additional cost.
The final utility function used for NBV optimization can be denoted as:
\begin{equation}
\begin{aligned}
\mathcal{U}(\mathbf{P})&=\mathcal{I}(\mathbf{P})-
w\mathcal{C}_{col}(o(\mathbf{P}^{\vee})),
\end{aligned}
\end{equation}
where $\mathbf{P}^{\vee}$ represents the position vector of transformation $\mathbf{P}$, $w$ is the tuning weight, and $\mathcal{C}_{col}(o)$ is the collision cost with gain $a$ and threshold $\tau$, which has the following format:
\begin{align}
    \mathcal{C}_{col}(o)=
    \left\{\begin{array}{lc}
      0
      &\text{for } o < \tau,\\
      e^{ao}-e^{a\tau}
      &\text{for } o \geq \tau.
    \end{array}\right.
    \label{eq:collision}
\end{align}

By maximizing the utility function $\mathcal{U}$, we can optimize the NBV pose and guide the robot's movement.

\section{Experiments}

\begin{table}[htpb]
    \caption{Variables and Their Default Values}
    \begin{center}
    \resizebox{0.45\textwidth}{!}{
    \begin{tabular}{ccl}
      \toprule 
      { \textbf{Parameters}  } 
      & { \textbf{Value} } 
      & \multicolumn{1}{c}{\textbf{Description}} \\
      \midrule
      $N_k$ & $48$ & Number of views sampled for initialization\\
      \midrule
      \multirow{2}{*}{$N_r$, $N_e$} & \multirow{2}{*}{$5000$} & Number of rays for reconstruction \\ & & and NBV evaluation\\
      \midrule
      \multirow{2}{*}{$N_s$, $N_f$, $N_p$} &\multirow{2}{*}{ $16$ }& Number of surface points, free points \\& & and NBV evaluation points\\
      \midrule
      $\sigma_d$ & $5mm$ & Variance of surface point sampling \\
      \midrule
      $\lambda_1, \lambda_2$ & 2, 0.5 & Reconstruction loss tuning weights\\
      \midrule
      \multirow{2}{*}{$w, a, \tau$} & \multirow{2}{*}{10, 1, 0.1} & Weight, gain and threshold of collision
      \\& & cost $\mathcal{C}_{col}(o)$ in Eqn. (\ref{eq:collision})\\
      \midrule
      $\mathbf{p}_{min}, \mathbf{p}_{max}$ & $\pm 0.25m$ & Bounding box of the object\\
      \bottomrule
    \end{tabular}
    }
  \end{center}
  \vspace{-6mm}
\label{tab:parameters}
\end{table}

\begin{figure*}[htpb]
    \centering
    \includegraphics[width=0.95\textwidth]{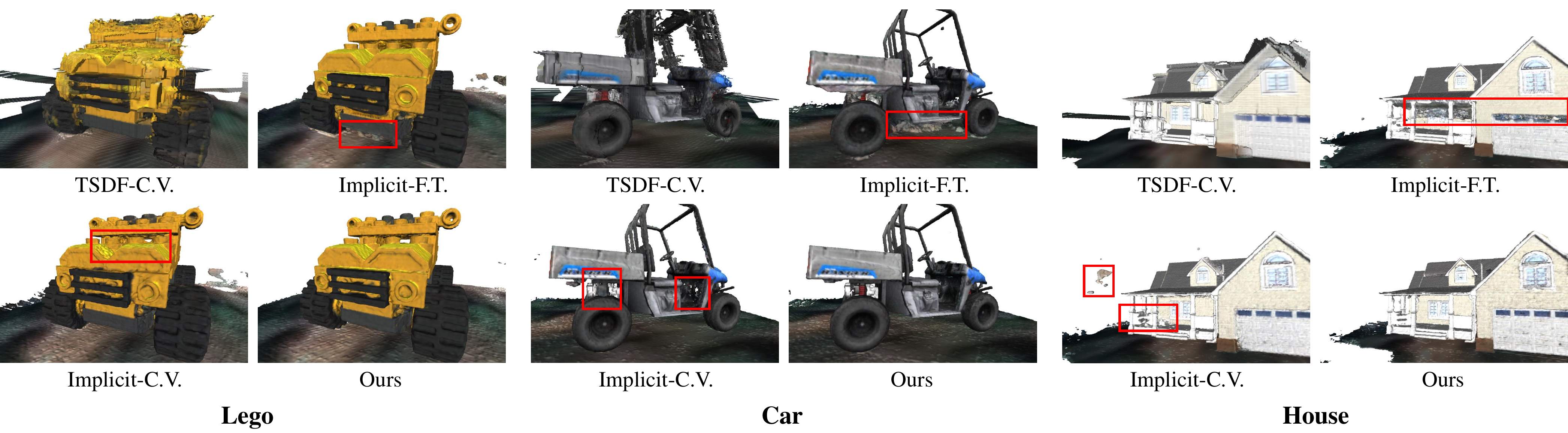}
    \caption{
      Qualitative comparison results of Lego, Car, and House.
      It is challenging to reconstruct complex, realistic objects using only ten views.
      Our method can detect uncertain parts of the object and assign views for optimal observation.
      It removes artifacts generated in fixed view settings and generates better results against the candidate view method.
      The implicit reconstruction pipeline also builds finer geometry than traditional TSDF fusion methods.
    }
    \label{fig:accuracy}
\end{figure*}

\subsection{Implementation Details}

\begin{table*}[!t]
    \tiny
    \centering
    \caption{
        Quantitative comparison results of Lego, Car, and House.
        We evaluate the \textit{Accuracy} ($cm$), \textit{Completion} ($cm$), \textit{CompletionRatio} (\%) metrics.
        The bold number represents the best result, and the underline represents the second best.
    }
    \vspace{-1mm}
    \label{tab:accuracy}
    \resizebox{0.85\textwidth}{!}
    {
        \begin{tabular}{ccccccccccc}
            \toprule
            \multicolumn{2}{c}{\multirow{2}*{\textbf{Methods}}} & \multicolumn{3}{c}{\textbf{Lego}} & \multicolumn{3}{c}{\textbf{Car}} & \multicolumn{3}{c}{\textbf{House}} \\
            \cmidrule{3-11}
            & & \textbf{Acc.}$\downarrow$ & \textbf{Comp.}$\downarrow$ & \textbf{C.R.}$\uparrow$ & \textbf{Acc.}$\downarrow$ & \textbf{Comp.}$\downarrow$ & \textbf{C.R.}$\uparrow$ & \textbf{Acc.}$\downarrow$ & \textbf{Comp.}$\downarrow$ & \textbf{C.R.}$\uparrow$ \\
            \midrule
            \multirow{3}*{\textbf{TSDF}} & \textbf{F.T.} & 1.073 & 1.390 & 41.043 & 1.244 & 0.472 & 84.472 & \underline{2.315} & 0.537 & 85.125\\
            & \textbf{R.V.} & \textbf{0.975} & 1.324 & 45.078 & 1.244 & 0.459 & 85.734 & \textbf{2.311} & 0.573 & 82.889\\
            & \textbf{C.V.} & 0.984 & 1.284 & 47.246 & 1.271 & 0.389 & 90.000 & 2.324 & 0.425 & 89.424\\
            \midrule
            \multirow{3}*{\textbf{Implicit}}  & \textbf{F.T.} & 2.005 & 0.911 & 66.124 & 1.244 & 0.243 & 95.849 & 2.694 & 0.274 & 97.506\\
            & \textbf{R.V.} & 2.024 & 0.941 & 63.234 & 1.266 & 0.287 & 93.486 & 2.591 & 0.289 & 96.217\\
            & \textbf{C.V.} & 1.956 & \underline{0.876} & \underline{68.050} & \textbf{1.087} & \underline{0.193} & \underline{99.014} & 2.643 & \textbf{0.249} & \underline{97.592}\\
            \midrule
            \multicolumn{2}{c}{\textbf{Ours}} & \underline{0.977} & \textbf{0.701} & \textbf{73.654} & \underline{1.215} & \textbf{0.185} & \textbf{99.197} & 2.682 & \underline{0.256} & \textbf{98.710}\\
            \bottomrule
        \end{tabular}
    }
    \vspace{-2mm}
\end{table*}

Our method runs on a mobile robot equipped with an RGB-D camera.
We implement our method using Pytorch and utilize ROS for robot driver and visualization.
We use tiny-cuda-nn\cite{tiny-cuda-nn} to boost our implicit network and Adam optimizer \cite{kingma2014adam} for the network training, pose refinement, and NBV pose optimization.
The learning rates of the optimizers are set to $2e^{-3}$, $3e^{-3}$, and $1e^{-2}$ respectively.
The settings of parameters mentioned in Section \ref{sec:recon} and Section \ref{sec:NBV} are shown in Table \ref{tab:parameters}.
We set the cosine scheduler for top-N strategy as $N_t = N_e \cos(\frac{\pi}{30}N_v)$.
We freeze the model after $200$ iterations of reconstruction and run another $100$ iterations to optimize the NBV pose.
During view pose optimization, we define the manifold $\mathit{\Psi}(n)$ as poses that face towards the center of the object.
For initialization, we randomly sample poses in a shell between radius $0.4m$ and $0.6m$ while keeping their occupancy smaller than $0.1$.
When the robot is guided to a new view, we add it to the reconstruction view set and increase $N_v$ by one.
Our reconstruction is complete after $10$ views are observed.

\subsection{Simulation Experiments}

\begin{figure}[htpb]
  \vspace{-2mm}
  \centering
  \begin{minipage}[b]{0.1125\textwidth}
    \centering
    \includegraphics[width=\textwidth]{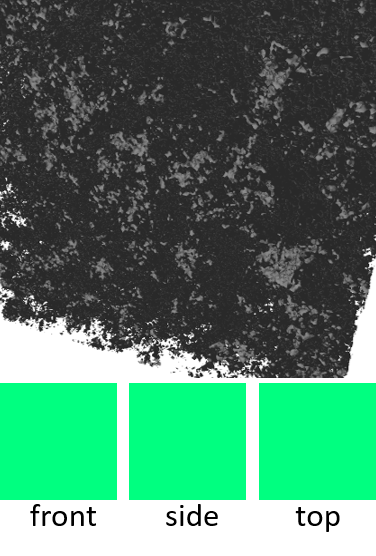}
    \footnotesize{(a) step = 0}
  \end{minipage}
  \begin{minipage}[b]{0.1125\textwidth}
    \centering
    \includegraphics[width=\textwidth]{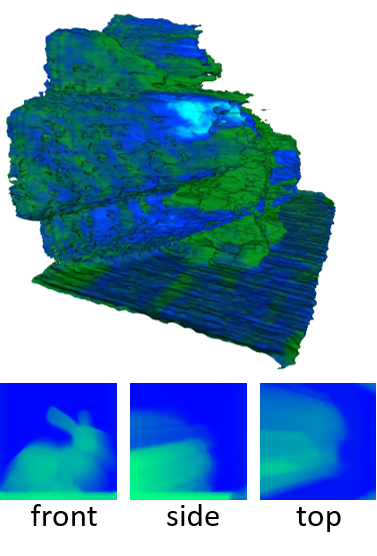}
    \footnotesize{(b) step = 1}
  \end{minipage}
  \begin{minipage}[b]{0.1125\textwidth}
    \centering
    \includegraphics[width=\textwidth]{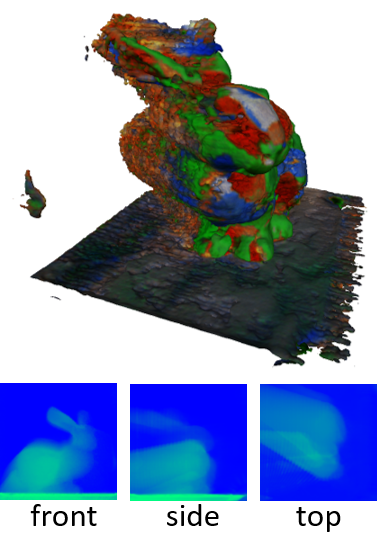}
    \footnotesize{(c) step = 2}
  \end{minipage}
  \begin{minipage}[b]{0.1125\textwidth}
    \centering
    \includegraphics[width=\textwidth]{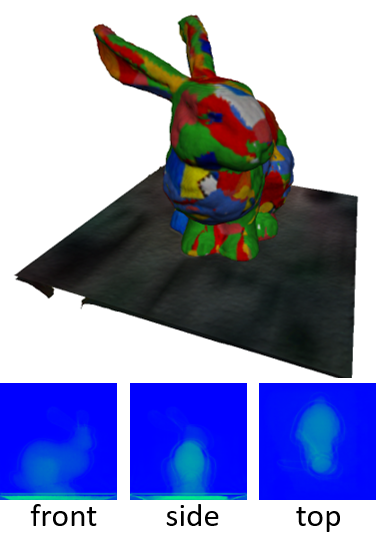}
    \footnotesize{(d) step = 20}
  \end{minipage}
    
  \vspace{-2mm}
  \caption{
    Visualization of the reconstruction process.
    The reconstruction results from steps 0, 1, 2, and 20 are shown in colored mesh extracted by Marching Cubes \cite{lorensen1987marching}.
    The bottom row shows the entropy map, viewed in a three-view perspective by sum projection.
  }
  \label{fig:recon_process}
\end{figure}

We use Gazebo as our simulation environment.
In our simulation scene, the target object is placed at the center of a room.
Color and depth images are captured by a flying RGB-D camera, the movement of which is controlled by our NBV planner.
After implicit reconstruction is finished, the Marching Cubes algorithm \cite{lorensen1987marching} is used to extract colored mesh representation for visualization.
Figure \ref{fig:recon_process} shows the reconstruction process from start to end.

\subsubsection{Reconstruction Process Evaluation}

In order to test the reconstruction efficiency of our reconstruction process, we conduct comparative experiments against traditional volume-based methods \cite{isler2016information} on three toy objects (Stanford Bunny, Dragon, and Armadillo).
We compare against various uncertainty policies including \textit{Occlusion-Aware} metrics, \textit{Average-Energy} metrics, as well as \textit{Random View} method.
We visualize the reconstructed object model and compare the surface coverage as a function of $N_v$.

We show the quantitative and qualitative evaluation results in Fig. \ref{fig:compare_result}.
In the reconstructed meshes, we can see that our implicit surface models have finer geometry and smoother surface than voxel representation.
By evaluating the curve of surface coverage, we find that our method achieves faster convergence and more efficient reconstruction than other methods.

\subsubsection{Final Model Evaluation}

To evaluate our method's final reconstruction accuracy and completion, we perform quantitative and qualitative evaluation on three realistic objects (Lego, Car, and House).
These objects are rescaled to fit our object bounding box.
We evaluate the reconstructed mesh against ground truth using \textit{Accuracy}, \textit{Completion}, \textit{CompletionRatio} metrics with the threshold of $0.01m$.
We compare our method with TSDF \cite{curless1996volumetric} methods and implicit methods based on our reconstruction pipeline using fixed circular trajectory (\textit{\textbf{F.T.}}), random views (\textit{\textbf{R.V.}}), and candidate views (\textit{\textbf{C.V.}}).

The final results are shown in Fig. \ref{fig:accuracy} and Table \ref{tab:accuracy}.
Compared with methods using fixed or random trajectories, our method can adaptively select views with large uncertainties and remove artifacts due to missed collections.
The candidate views method achieves informative planning but still can not reach optimal solution.
Our method also obtains higher-precision reconstruction results against the TSDF method, leveraging its implicit surface and specially designed pipeline.

\subsubsection{Ablation Studies}

\begin{figure}[htpb]
    \vspace{-4mm}
    \centering
    \includegraphics[width=0.36\textwidth]{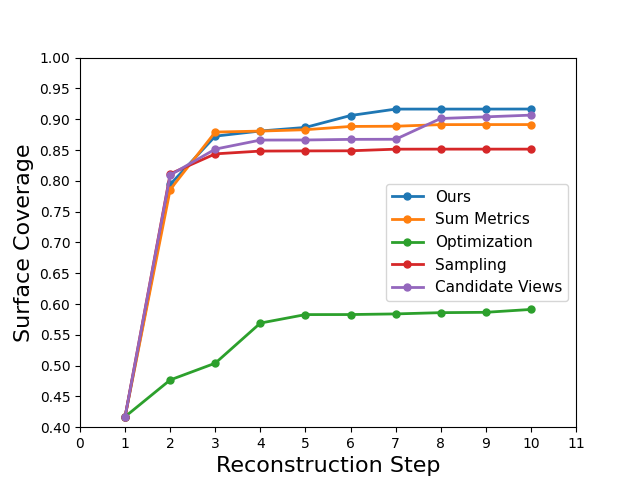}
    \caption{
      Comparison study of the NBV strategies.
      \textit{Sum Metrics} uses the simple information gain in Eqn. (\ref{equ:sum});
      \textit{Optimization} removes NBV sampling initialization;
      \textit{Sampling} directly uses randomly sampled views;
      \textit{Candidate Views} evaluates from uniformly distributed candidates.
      With the help of top-N evaluation metrics, optimization-based NBV planning and sampling initialization, our method escapes the local maximum and results in better reconstruction and faster convergence than other methods.
    }
    \label{fig:strategies}
\end{figure}

We conduct ablation studies to demonstrate the effect of our top-N metrics and NBV planning strategies.
In Fig. \ref{fig:strategies}, we hold our reconstruction pipeline constant and show the experiment results using different evaluation metrics and NBV planning strategies.
The results show that using sum metrics performs weakly in the later process due to the lack of attention to detail.
The optimization-based method using simple initialization got stuck in the local maximum.
With top-N metrics and sampling initialization, our method can refine details and avoid local maximum.
Leveraging the NBV optimization, our method also outperforms selection-based methods using preset candidate views, which can only obtain sub-optimal results.

\begin{figure}[htpb]
    \centering
    \begin{minipage}[b]{0.18\textwidth}
    \centering
      \includegraphics[width=\textwidth]{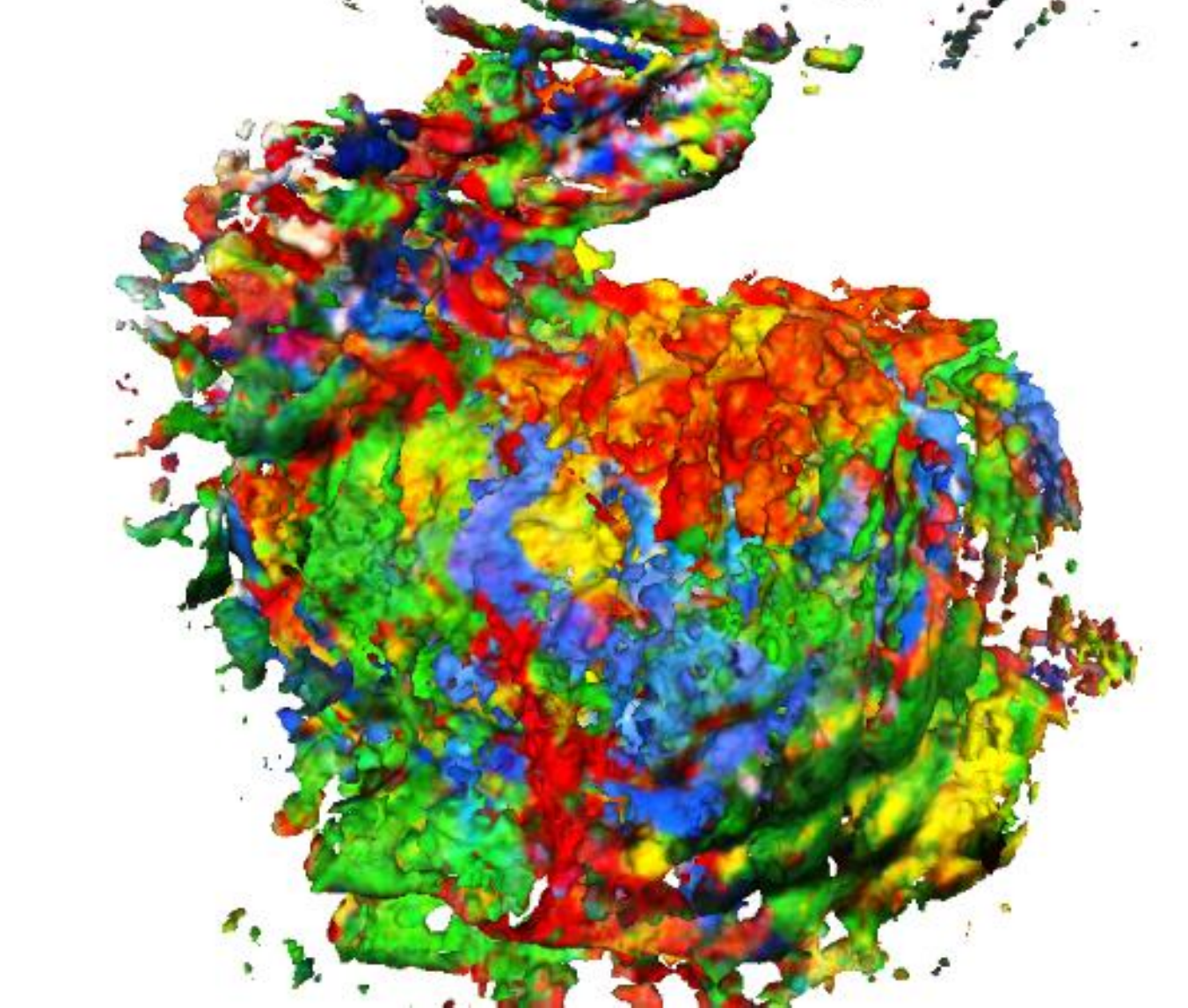}
      \footnotesize{(a) w/o pose refinement}
    \end{minipage}
    \begin{minipage}[b]{0.18\textwidth}
    \centering
      \includegraphics[width=\textwidth]{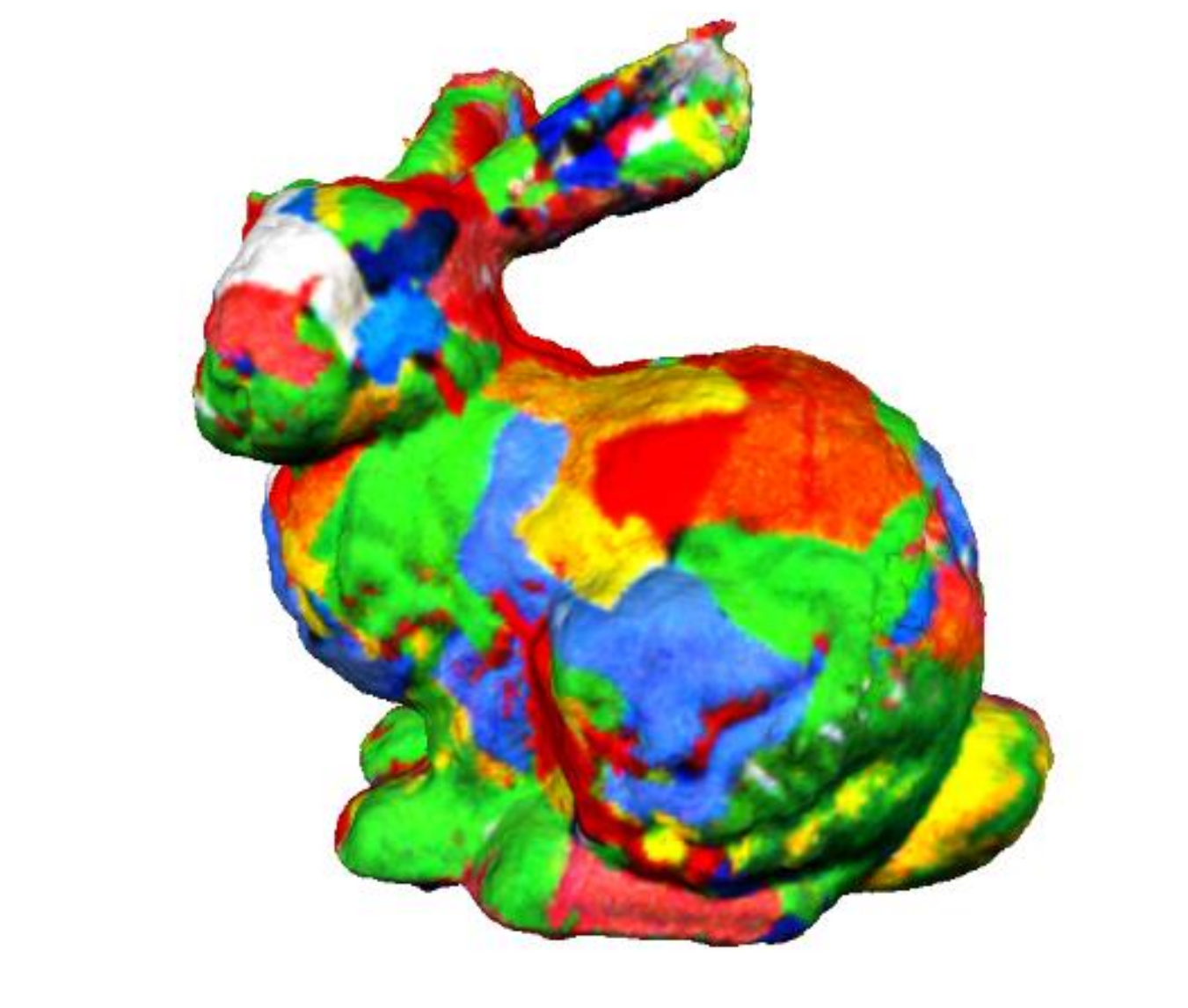}
      \footnotesize{(b) with pose refinement}
    \end{minipage}
    \caption{
        Ablation study on the use of pose refinement.
        The left shows the original noisy output, and the right shows the consistent model with pose refinement.
        Using multi-view information, noisy poses are optimized together with the implicit model to output better reconstruction.
    }
    \label{fig:pose_opt}
\end{figure}

We also evaluate the pose refinement module to demonstrate its adaptability to inaccurate reference poses.
In this experiment, we add noise to the ground truth poses in the simulation environment.
Specifically, we add uniform distribution of $U(-0.05, 0.05)$ radius to rotation, and $U(-0.05, 0.05)$ meter to translation.
The reconstruction results are shown in Fig. \ref{fig:pose_opt}.
In the visualization, we prove the ability of our pose refinement module to recover a consistent object model even despite severe frame misalignment.

\begin{figure}[htpb]
    \vspace{-4mm}
    \centering
    \begin{minipage}[b]{0.19\textwidth}
    \centering
        \includegraphics[width=\textwidth]{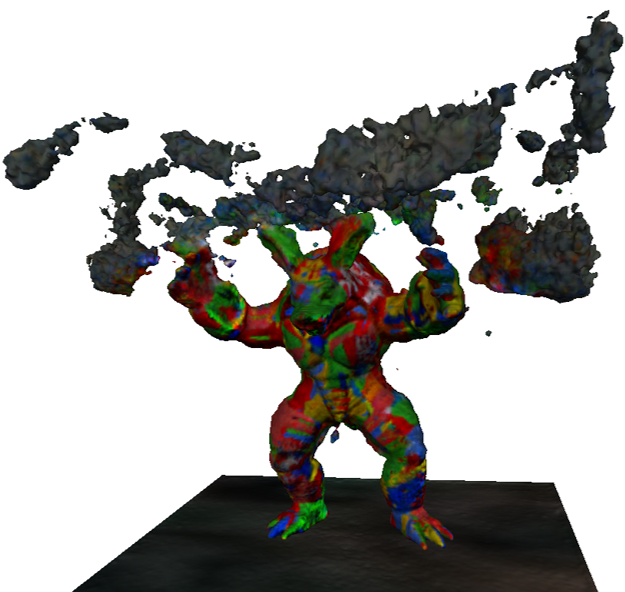}
        \footnotesize{(a) w/o free ray supervision}
    \end{minipage}
    \begin{minipage}[b]{0.19\textwidth}
    \centering
        \includegraphics[width=\textwidth]{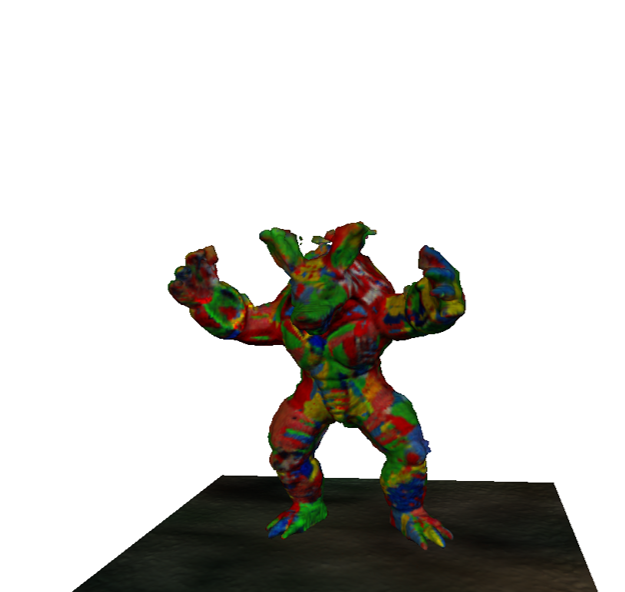}
        \footnotesize{(b) with free ray supervision}
    \end{minipage}
    \caption{
      Ablation study on the use of free ray supervision.
      With free ray supervision, rays with no valid depth in the depth map can be utilized, and floaters in these regions can be removed.
    }
    \vspace{-4mm}
    \label{fig:void_ablation}
\end{figure}

In Fig. \ref{fig:void_ablation}, we compare reconstruction with and without free ray supervision.
Free ray supervision not only removes backspace floaters and creates a cleaner model but also helps fully merge the observed information, which is helpful to subsequent NBV planning.

\subsection{Real-world Experiments}

\begin{figure}[htbp]
    \centering
    \vspace{-2mm}
    \includegraphics[width=0.40\textwidth]{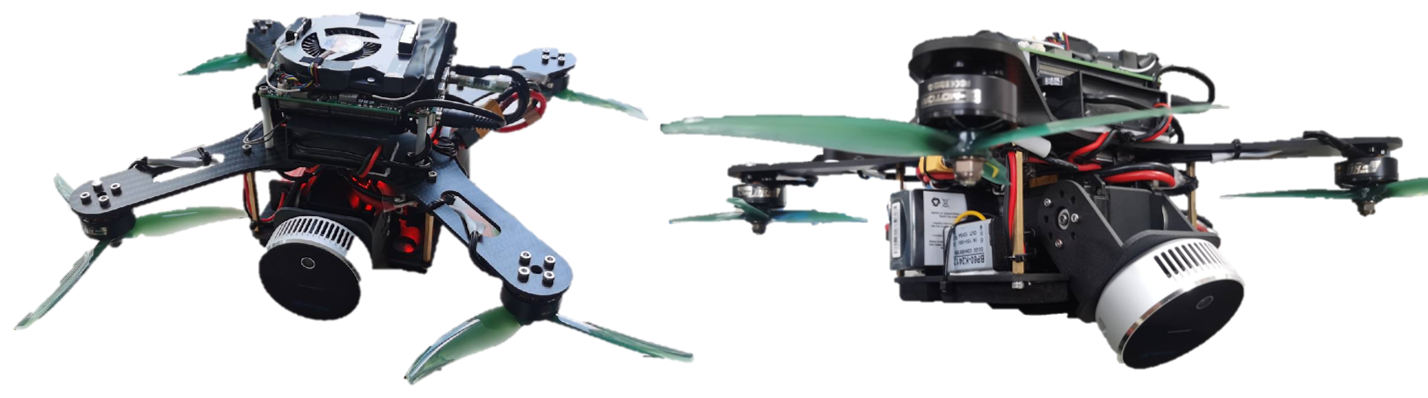}
    \vspace{-2mm}
    \caption{
      Our compact MAV equipped with tiltable lidar camera.
    }
    \label{fig:uav}
\end{figure}

In our real-world active reconstruction experiments, we use a compact micro aerial vehicle (MAV) equipped with a tiltable RGB-D camera (Intel Realsense L515) for data collection (Fig. \ref{fig:uav}) and a ground station with a single NVIDIA 3070ti GPU for computation.
Color and aligned depth images are streamed from the MAV to the computer to perform implicit reconstruction and NBV planning.
We use the low-cost ArUco markers \cite{garrido2014automatic} to estimate the pose of the sensor. 

\subsubsection{Qualitative Evaluation}

\begin{figure*}[htpb]
    \centering
    \includegraphics[width=0.95\textwidth]{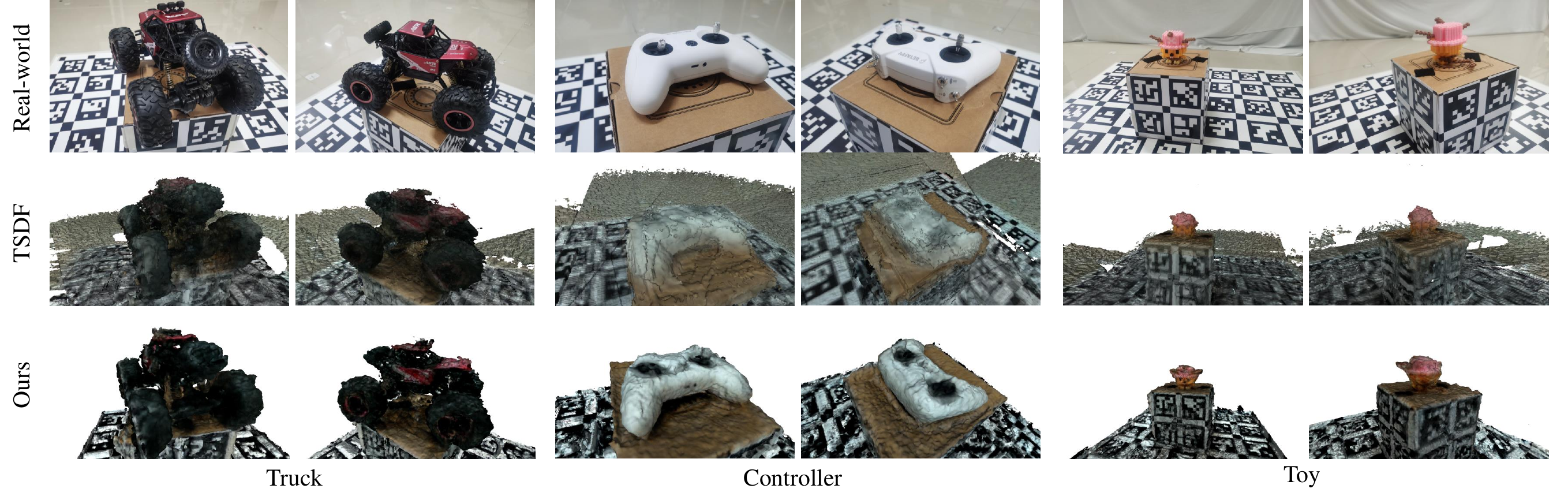}
    \caption{
     Comparison of our method against traditional TSDF method in real-world active reconstruction experiments of Truck, Controller, and Toy.
     Our method achieves robust active planning and fine reconstruction under challenging real-world settings like noisy depth input and inaccurate poses.
    }
    \vspace{-2mm}
    \label{fig:realworld_recon}
\end{figure*}

In Fig.~\ref{fig:realworld_recon}, we present the result of the real-world experiments.
We conduct active reconstruction on three real-world objects (Truck, Controller, and Toy) and compare our method against the TSDF method.
The results show that our method is feasible, robust and can generate precise and complete object models under challenging real-world settings.

\subsubsection{Time Consumption}

We perform time consumption evaluation of our method for each stage in the reconstruction process. The result is \textbf{\textit{(a)}} fusion of newly collected information: $4.81s$, \textbf{\textit{(b)}} optimization-based planning: $1.35s$ and 
\textbf{\textit{(c)}} robot movement (averagely measured using videos of real-world experiments): $15.20s$.
Our method achieves fast model training and NBV searching, introducing little additional overhead to the reconstruction process.

\section{Conclusion}

This paper explores the problem of active implicit reconstruction of unknown objects using implicit representation.
We build up an implicit occupancy field for object-level reconstruction and we evaluate view uncertainty directly from the occupancy probability using a sampling-based method.
Leveraging the differentiability of the implicit representation, we can do away with the candidate views and propose optimizing NBV by maximizing the view information.
Our method generates high-precision object model in both simulation and real-world experiments and achieves efficiently and intelligently during view planning.
It outperforms baseline methods in both model quality and feasibility.
In our future work, we aim to boost robot exploration tasks in more complex scenes by leveraging the optimization-based view planning method.



{
\bibliographystyle{IEEEtran}
\balance
\bibliography{reference}
}

\end{document}